\documentclass[runningheads]{llncs}
\usepackage[T1]{fontenc}
\usepackage{graphicx}
\usepackage{acronym}
\usepackage{enumitem}
\usepackage{float}
\usepackage{amsmath}

\usepackage{booktabs}
\usepackage{subcaption}
\usepackage{url}

\begin{document}
\title{A Multimodal Framework for DeepFake Detection}
%
%
\author{Kashish Gandhi\inst{1,2}\and
Prutha Kulkarni\inst{1,3}\thanks{Equal authorship}\and
Taran Shah\inst{1,4}*\and
Piyush Chaudhari\inst{1,5}*\and
Meera Narvekar\inst{1,6}\and
Kranti Ghag\inst{1,7}
}

\authorrunning{K. Gandhi et al.}

\institute{Department of Computer Engineering, Dwarkadas J. Sanghvi College of Engineering, Mumbai, India\\ \and
kashishgandhi6112003@gmail.com\\ \and
kulkarniprutha1@gmail.com\\ \and taran.shah9@gmail.com\\ \and piyush300504@gmail.com \\ \and
meera.narvekar@djsce.ac.in \\ \and
kranti.ghag@djsce.ac.in
}

\maketitle              
\begin{abstract}
The rapid advancement of deepfake technology poses a significant threat to digital media integrity. Deepfakes, synthetic media created using AI, can convincingly alter videos and audio to misrepresent reality. This creates risks of misinformation, fraud, and severe implications for personal privacy and security. Our research addresses the critical issue of deepfakes through an innovative multimodal approach, targeting both visual and auditory elements. This comprehensive strategy recognizes that human perception integrates multiple sensory inputs, particularly visual and auditory information, to form a complete understanding of media content. For visual analysis, a model that employs advanced feature extraction techniques was developed, extracting nine distinct facial characteristics and then applying various machine learning and deep learning models. For auditory analysis, our model leverages mel-spectrogram analysis for feature extraction and then applies various machine learning and deep learning models. To achieve a combined analysis, real and deepfake audio in the original dataset were swapped for testing purposes and ensured balanced samples. Using our proposed models for video and audio classification i.e. Artificial Neural Network and VGG19, the overall sample is classified as deepfake if either component is identified as such. Our multimodal framework combines visual and auditory analyses, yielding an accuracy of 94\%.

\end{abstract}

\section{Introduction}
Recent advances in deep learning and media technologies have made synthetic generation of media more accessible than ever. It requires minimal effort to allow consumers to manipulate all kinds of media and spread misinformation. This can lead to fraudulent activities, scams, and spreading of misinformation \cite{b1}, \cite{b2}. 


Several approaches have been proposed gravitating towards a unimodal approach, meaning they take only one modality into account, either audio or video \cite{b3},\cite{b4}. However, these unimodal approaches fall short in addressing complexity and nuance of sophistiacted deepfakes. Although these detectors have shown impressive performances, video-only detectors can be deceived by synthetics audios and vice-versa. Hence, these prove to be ineffective for robust deepfake detection. To overcome these limitations, a multimodal analysis framework is proposed, integrating visual, auditory, and textual features to provide a holistic view of the media content. This approach not only enhances detection capabilities but also offers a more resilient solution against the evolving landscape of deepfake technologies.

Through this paper, a new method that combines audio-visual information is presented. Custom features from the video dataset and spectrograms from the audio were extracted dataset during the training phase. Our models were trained on unimodal datasets which helps the developed detector to not overfit on a single data type and generalize as much as possible. 

\section{Literature Review}

Deepfake technology has become increasingly prevalent, with various tools available online for creating synthetic videos. Popular applications such as FakeApp, Faceswap, DeepFaceLab, and Face Swap Generative Adversarial Network utilize autoencoder-decoder and Generative Adversarial Network architectures to produce realistic deepfakes.  In audio deepfakes, Google's text-to-speech technology has driven a rise in usage, supported by tools like WaveNet by DeepMind and TacoTron by Google as mentioned in the survey \cite{b5}. These technologies often use Variational Autoencoders (VAEs), which compress and reconstruct audio to mimic target speakers, and Generative Adversarial Networks (GANs) for audio manipulation. This proliferation of deepfake technology presents significant cybersecurity threats, prompting research into effective detection methods.


Unimodal approaches to deepfake detection—those focusing solely on either audio or visual analysis—often fall short in handling the complexities of sophisticated deepfakes. Studies below by various researchers demonstrate that relying solely on visual features, such as facial characteristics or eye blinking patterns, can lead to inaccuracies as these methods may be vulnerable to variations in data or specific attack vectors that manipulate only one modality. A. Ismail et al. \cite{b6} used FaceForensics and CelebDF datasets with InceptionResNetV2 for feature extraction, YOLO for face detection, and XGBoost for classification, achieving 90\% accuracy. However, their approach's reliance on fixed feature extraction limits its effectiveness on new data. Similarly, \cite{b7} tracked eye blinking patterns to differentiate real and deepfake videos using Fast-Hyperface, but this method is sensitive to individual blinking variations and environmental factors, leading to false positives or missed detections. Our method will mitigate this by combining more than one visual features, providing a more comprehensive detection framework that is less reliant on a single feature. 

Computational limitations and the use of incompatible models can significantly impact the effectiveness of deepfake detection systems. Many existing studies, such as those by \cite{b8},\cite{b9} and \cite{b10}, highlight how the choice of models and computational resources can constrain the performance of detection algorithms. For instance, \cite{b8} combined resource-intensive models like ConvNeXt, Swin Transformer, Autoencoder, and Variational Autoencoder, requiring substantial computational power, which is a barrier for real-time applications. Additionally, \cite{b9} used a 3D CNN to process videos directly, capturing spatial and temporal information but leading to high computational costs and significant memory use.  Concurrently, in the study \cite{b10} they have used 7 different models including CapsuleForensics and Xception achieving up to 77\% accuracy. Even with a custom-curated dataset, their approach only achieved limited success due to the use of multiple models that were not well-integrated with the type of data they analyzed, leading to inefficiencies and reduced accuracy.


Our research journey in deepfake detection was significantly shaped by insights from various key papers. The study began by exploring \cite{b11}, which utilized MTCNN for face detection and EfficientNet-B5 for feature extraction, illustrating the importance of accurate face detection in our pipeline. This led us to integrate reliable face detection with advanced feature extraction. Considering \cite{b12}, which employed a 3D CNN to capture spatial-temporal features directly from video frames, highlighting the value of capturing video frames to reduce the processing time and computational resources. Inspired by \cite{b13}, which combined CNN-generated features with LSTM for sequential analysis, recognizing the importance of integrating sequential models for detecting long-term dependencies in videos. Also after reviewing \cite{b14}, which used CNN architectures like VGG19 for detecting facial artifacts, guiding us to evaluate and fine-tune these models. Furthermore, \cite{b15} emphasized the critical role of facial features such as eyes and mouth, leading us to focus on feature extraction methods sensitive to these regions. Additionally, \cite{b16} demonstrated the benefits of Convolutional Neural Networks (CNNs) to capture spatial-temporal dynamics and extract detailed features, inspiring us to apply CNN to for deepfake detection.

In recent research, several advancements have been made in the field of audio processing and deepfake detection. \cite{b17} explored deepfake audio detection by leveraging Explainable Artificial Intelligence (XAI). Their study emphasized the need for interpretability in deepfake detection systems, which is crucial for understanding and justifying the model’s predictions. Study in \cite{b18} compared Mel Frequency Cepstral Coefficients (MFCC) and Mel spectrograms for raga classification using CNNs. Their comparative analysis provided insights into which feature extraction technique yields better results for audio classification tasks. They found that while both MFCC and Mel spectrograms are effective, Mel spectrograms offer more detailed temporal and spectral information, which can be advantageous for tasks like raga classification and potentially for deepfake audio detection as well. 

A more advanced method for creating mel-spectrograms involved converting MFCC to mel-spectrograms. \cite{b19} implemented a custom 7-layer CNN model but couldn't determine its real-world performance. Observing that tokenizing words for classification was challenging as identical words were spoken in human and fake voices. This led us to focus on pitch, a key differentiating factor, and thus adopted mel-spectrograms for better pitch representation. The work by \cite{b20} applied MFCC feature extraction and classified audio samples using SVM and VGG-16 transfer learning technology. Additionally, \cite{b21} demonstrated that transfer learning with the VGG19 model, a pre-trained 19-layer network augmented with additional layers, effectively classifies sound images and their corresponding audio. Thus, inspired by these approaches, VGG19 technology and mel-spectograms for audio analysis were explored. Recent studies have demonstrated the potential of multimodal methods in improving the detection of synthetic media. For instance in \cite{b22},\cite{b23} and \cite{b24}, the authors used a multimodal approach, extracting and fusing facial and speech features to improve deepfake detection accuracy, showing that multimodal systems can outperform unimodal counterparts in detecting deepfakes, inspiring us to adopt a similar strategy.


By integrating these insights, a deepfake detection system that addresses the limitations of unimodal approaches by combining both visual and audio features was developed. Our system is designed to be computationally efficient, leveraging advanced feature extraction and sequential models, and is capable of generalizing across different datasets and deepfake types. This comprehensive approach ensures robust and accurate detection, overcoming the challenges identified in previous research.

\section{Data Collection}

\subsection{Video Dataset}
 A dataset that matched the characteristics of contemporary Deepfake videos was selected. The training subset, derived from DFDC Dataset, \cite{b25} consists of 2,619 deepfake videos and 515 real videos, representing different age, race, gender, and demographic groups, ensuring a broad and inclusive training dataset. This diversity is important very to develop a robust model that can accurately detect deepfake in different populations segments.

\subsection{Audio Dataset}
For audio dataset, the Fake-or-Real (FoR) dataset \cite{b26} was utilized, which comprises over 195,000 utterances from both real human speech and computer-generated synthetic speech. The dataset is available in four versions: for-original, for-norm, for-2sec, and for-rerec. The for-norm version, containing 26,927 fake and 26,941 real audio samples, and the for-rerec version, which includes an equal split of 5,104 fake and 5,104 real samples was selected. Our model was trained on a subset of 4,000 samples from this dataset ranging from 2 to 4 seconds in duration, providing a diverse range of audio for effective detection of synthetic speech patterns.

\section{Proposed Methodology}
 The proposed methodology includes a multi modal approach to make one of 4 classifications whether it is:\\
 a) Real video with Real audio\\
 b) Real video with Deepfake audio\\
 c) Deepfake video with Real audio \\
 d) Deepfake video with Deepfake audio.\\ The input video will first have its audio extracted and passed through our audio deep fake detection model which generates and extracts the perfect Mel-Spectrogram of the audio samples as well as MFCC \cite{b27} for the same. These were passed through various models for classification as real and deep fake audio. The video on the other hand was passed through the feature extraction model which was stored as an array and consequently passed through the classification model. This resulted in an output from the mentioned outcomes

\begin{figure}[H]
    \centering
    \includegraphics[width=12cm]{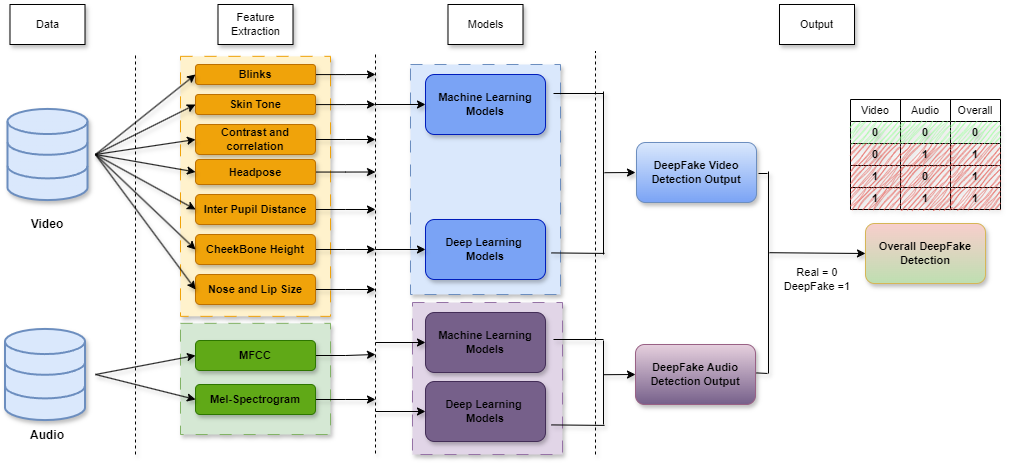}
    \caption{\label{fig:equation1}Pipeline for proposed methodology}
    \end{figure}

\subsection{Feature Extraction}
\subsubsection{4.1.1 For Video}
As the initial step in preprocessing our data, nine features were extracted from the videos. Seven of these features involved detecting facial landmarks and extracting pertinent attributes. To achieve this, the Haar Cascade algorithm, since it can easily detect objects in images irrespective of their scale and location. Hence, it was employed to accurately identify faces within the frame, defining our region of interest (ROI). Subsequently, FaceMesh’s Face Landmark Model was utilized to detect specific features within the identified ROI:
\begin{itemize}
    \item \textbf{Nose and Lip Size:}
    Research shows in \cite{b28} that size of facial features is often tampered with when a deepfake is generated. Hence, it was important to analyze the same for real and deepfake videos. Landmark indices used were 1 and 197 for the base and tip of the nose and 61 and 291 for the left and right corners of the nose, respectively, obtained using facemesh. The distance between the two points was found using the distance formula. A similar approach was used to calculate the lip size where the distance between either corner of the lips was found and computed using the euclidean distance formula.
    \begin{figure}[H]
        \centering
        \begin{subfigure}{0.3\textwidth}
            \includegraphics[width=\linewidth, height=4.5cm, width=4cm]{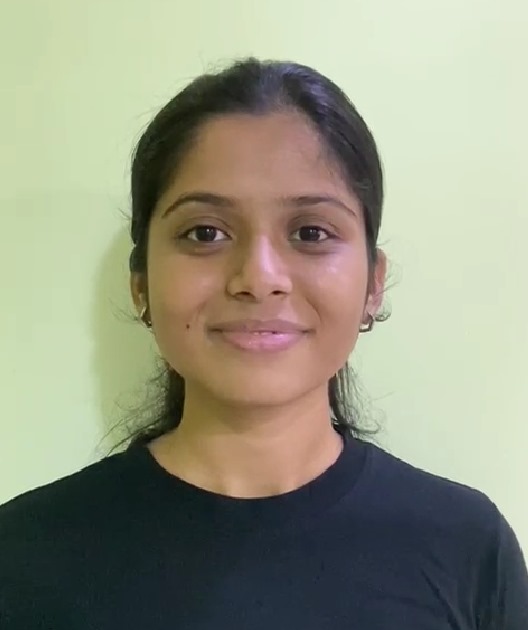}
            \caption{}
        \end{subfigure}
        \hfill
        \begin{subfigure}{0.3\textwidth}
            \includegraphics[width=\linewidth, height=4.5cm, width=4cm]{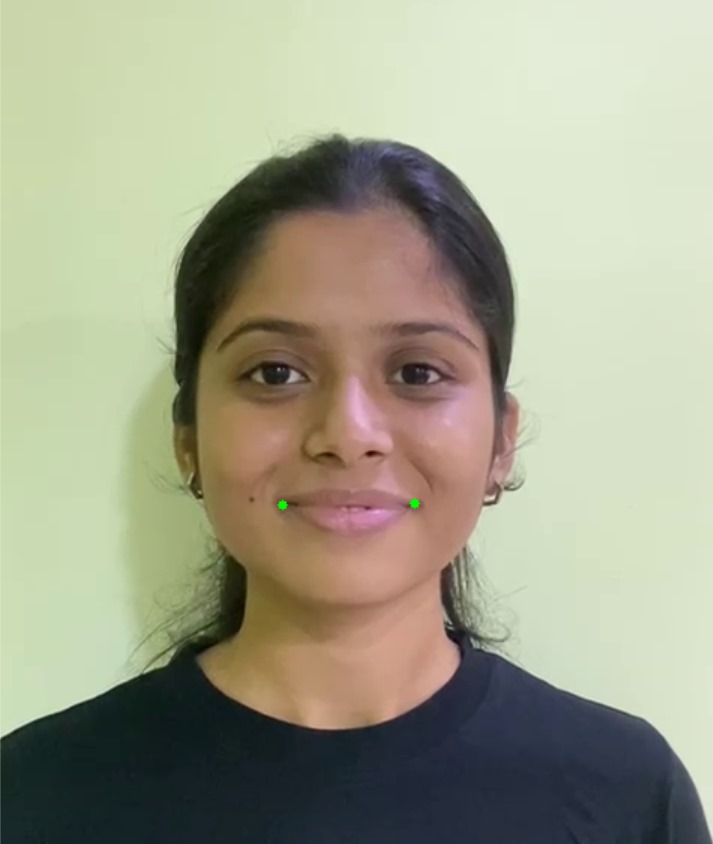}
            \caption{}
        \end{subfigure}
        \hfill
        \begin{subfigure}{0.3\textwidth}
            \includegraphics[width=\linewidth, height=4.5cm, width=4cm]{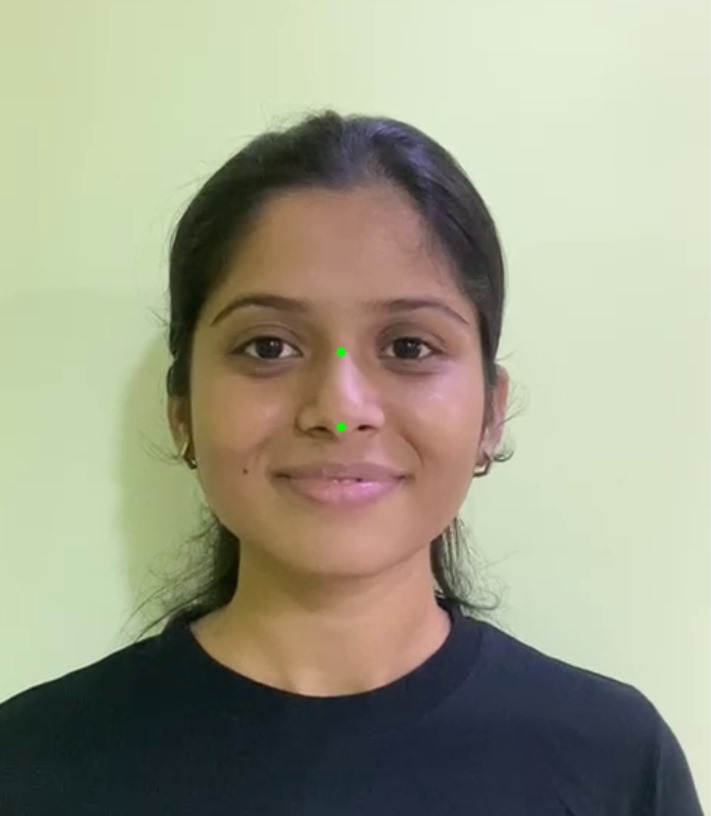}
            \caption{}
        \end{subfigure}
        \caption{The image shows the process of detecting deepfake alterations in facial features by analyzing the distances between key nose and lip landmarks. Subfigures are labeled as follows (left-to-right): (a) Original image, (b) Lip Indices, and (c) Nose Indices.}
        \label{fig:multi_figs}
    \end{figure}

    \item \textbf{Contrast and Correlation:}
    Textural features are complex features that indicate roughness and regularity of an image. Textural features based on the Gray Level Co-Occurrence Matrix were extracted. It describes texture using the spatial distribution of pixels in an image. Using this matrix, the probability of two gray pixels being adjacent is found by computing distance and direction. The probability on different gray levels constitutes the gray level co-occurrence matrix. According to a previous study by Xu, Bozhi et al in \cite{b29}, a total of 14 features can be derived from the GLCM. In our study, two features from this matrix: contrast and correlation were selected since they help extract most valuable spatial relationships between pixels.

Contrast measures the richness and depth of texture details, with higher values indicating a greater gray-scale difference between pixels. The formula for calculating contrast is as follows:

    \begin{equation}
        f_{\text{Con}} = \sum_{i,j=1}^{N} P_{i,j} (i - j)^2
    \end{equation}

    Correlation measures the degree of correlation between elements of the gray level co-occurrence matrix.
    \begin{equation}
        f_{\text{Cor}} = \frac{\sum\limits_{i,j=1}^{N} (i - \mu_i) (j - \sigma_j) P_{i,j}}{\sigma_i \sigma_j}
    \end{equation}

    where
    \begin{equation}
        \mu_i = \sum\limits_{i,j=1}^{N} i P_{i,j}
    \end{equation}

    \begin{equation}
        \mu_j = \sum\limits_{i,j=1}^{N} j P_{i,j}
    \end{equation}

    \begin{equation}
        \sigma_i = \sqrt{\sum\limits_{i,j=1}^{N} P_{i,j} (i - \mu_i)^2}
    \end{equation}

    \begin{equation}
        \sigma_j = \sqrt{\sum\limits_{i,j=1}^{N} P_{i,j} (j - \mu_j)^2}
    \end{equation}

    N is the size of the gray-level co-occurrence matrix, and \( P_{i,j} \) is the value of the i-th row and j-th column of the gray-level co-occurrence matrix.
    To implement the above. This ROI was divided into 9 blocks to effectively deal with areas of the face that had been tampered with. Textural features were accurately extracted from each of the 9 sub-blocks and stored.\\

    \item \textbf{Blinks:}
    As mentioned in the research \cite{b7} by T. Jung et al. often it is seen that deepfakes have an irregular pattern in the blinking of the eye. This method tracks the blinking of the eyes as one of the features used in determining whether the video is real or deepfake. A combination of 2 models to extract the eye blinking count during the duration of a video was used. The outline of lining of the eye was noted, which helped find a ratio between the vertical height between the eyelids and the horizontal distance between the opposite corners of the eyes. The landmarks marked: 22, 23, 24, 26, 110, 157, 158, 159, 160, 161, 130, 243 were used to delineate the lining of the eye, as shown in the Fig 3.
    \begin{figure}[H]
    \centering
    \includegraphics[width=4cm, height=4cm]{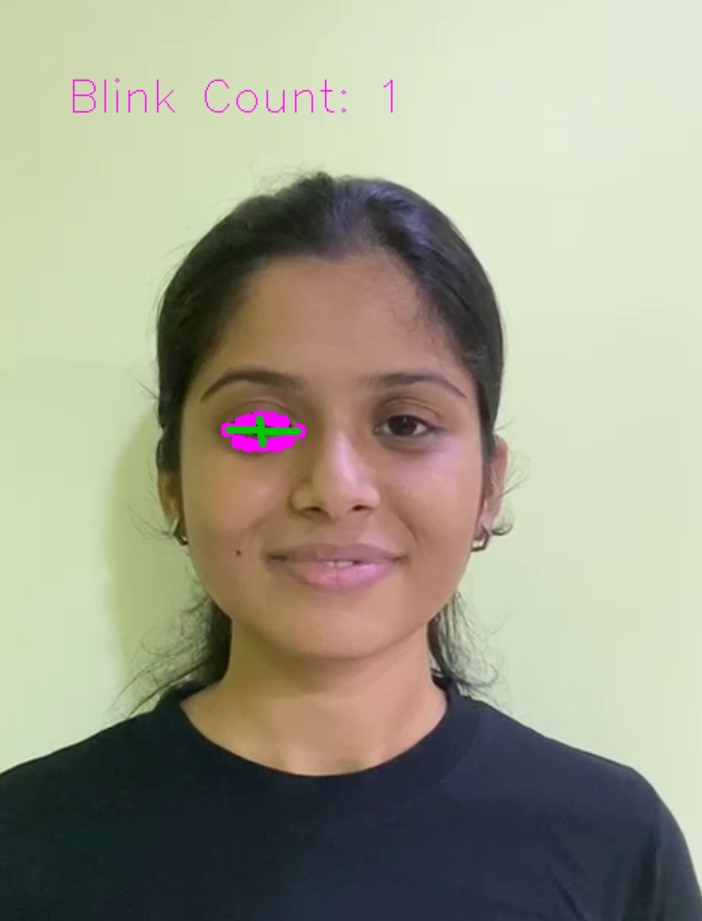}
    \caption{\label{fig:equation1}Blink feature extraction by marking key eye landmarks and tracking blinking patterns.}
    \end{figure}

    \item \textbf{Inter Pupil Distance:}
    Often when deepfakes are created, the normal inter-pupil distance is tampered with, making it a notable feature that can be used to distinguish between real and deepfake videos. The landmarks at the center top and center bottom of each eye were noted. Using each of those distances and taking their average ,a very accurate estimate of the inter-pupil distance was found. 
    \begin{figure}[H]
    \centering
    \includegraphics[width=9cm]{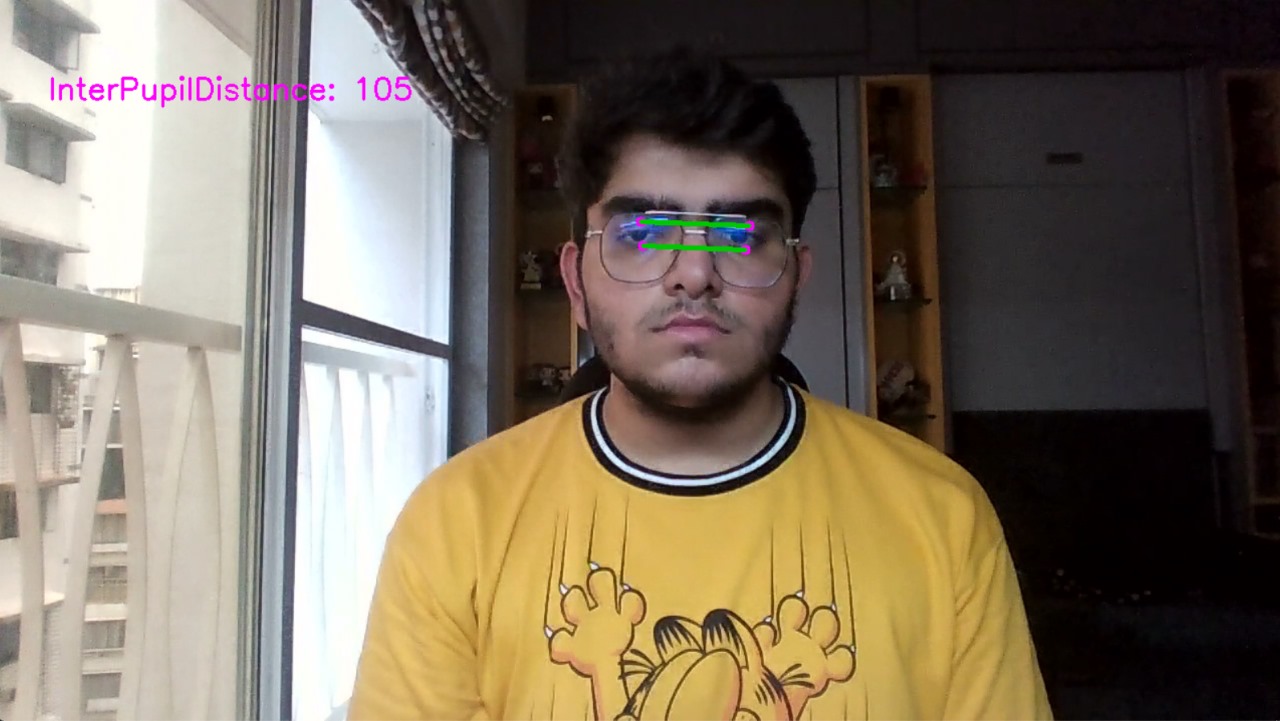}
    \caption{\label{fig:equation1}Inter-pupil distance feature extraction, using landmarks at the center top and bottom of each eye to accurately measure the distance.}
    \end{figure}

    \item \textbf{Cheek Bone Height:}
    Cheekbone height helps analyze whether the video is a deepfake or real as abnormal distance from the chin to the cheekbones is a common occurrence in deepfake videos. For calculating the height from the chin to the cheekbones, use of basic geometry was made. A combination of the sine and cosine rules was used to find the cheekbone height. Figure 5 shows an outline of the skeleton used to mark the cheekbone height, along with the equations used to determine the actual cheekbone height.
      \begin{figure}[H]
    \centering
    \includegraphics[width=10cm]{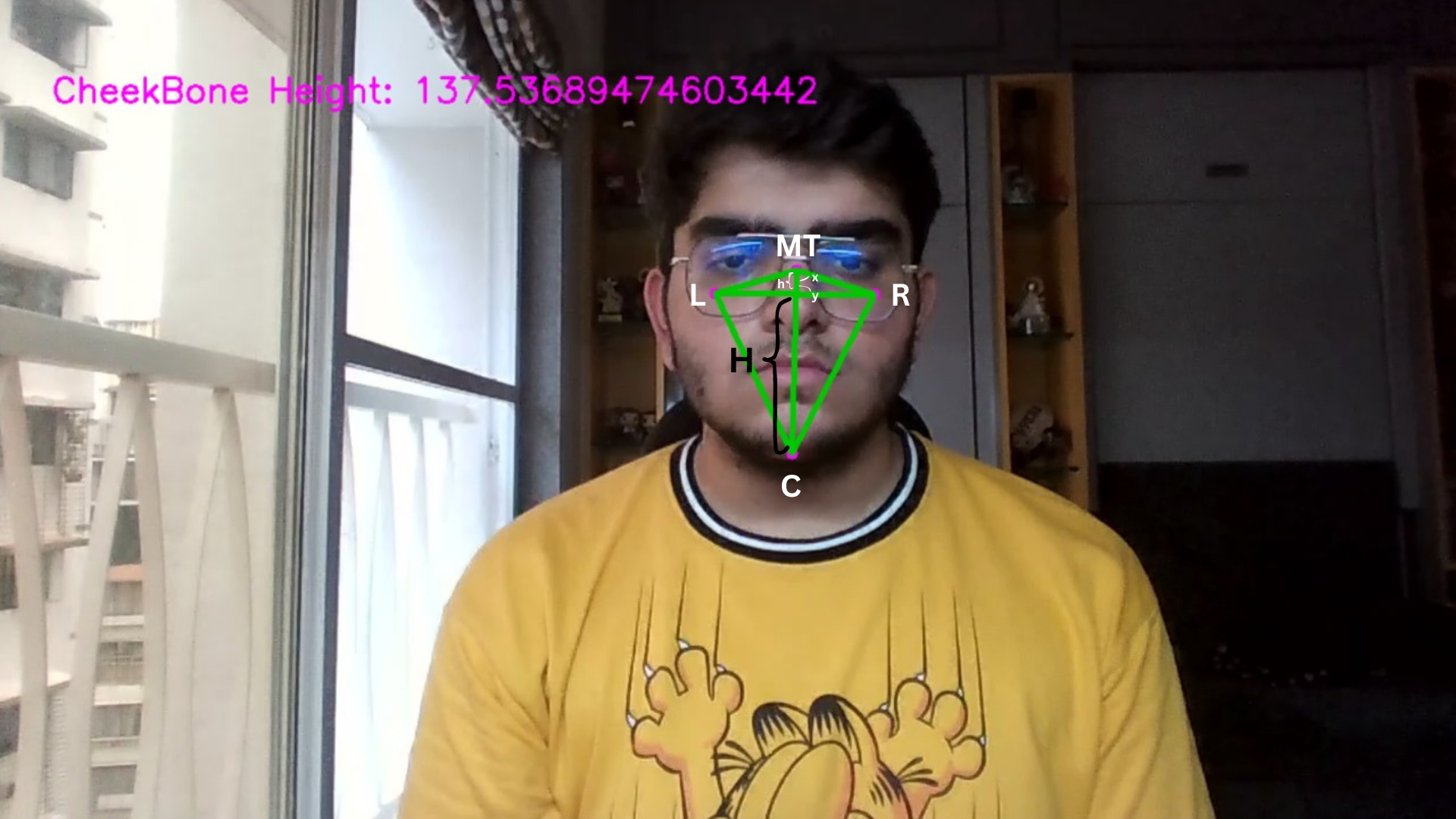}
    \caption{\label{fig:equation1}The extraction of cheekbone height feature, including the diagram used for implementing the mathematical calculations.}
    \end{figure}
     Our proposed analytical calculations:
    \begin{equation}
        \cos R = \frac{LR^2 + MTR^2 - MTC^2}{2 \cdot LR \cdot MTR}
    \end{equation}

    \begin{equation}
        \cos x = \frac{ MTR^2 + MTC^2 - RC^2}{ 2 \cdot MTR \cdot MTCR}
    \end{equation}

    \begin{equation}
        y = 180^\circ - (x + R)
    \end{equation}

    \begin{equation}
        \frac{\sin R}{h} = \frac{\sin y}{MTR}
    \end{equation}

    \begin{equation}
        h = \frac{\sin R \cdot MTR}{\sin y}
    \end{equation}

    \begin{equation}
        H=MTC-h
    \end{equation}

    Where: LR is the horizontal distance between the two cheekbones, MTC is the distance between the middle of the nose and the chin. R is the angle between MTR and LR, x is the angle between MTR and MTC and H is the height difference between the cheekbones and the chin.\\
    To implement the above. The face region was divided into 4 sections using the quadrilateral kite. The cosine rule is used to find out angle R and angle x. These are the used to find angle y. Further, sine rule is used to find h and subtracting h with MTC results in H which is the cheekbone height. \\

    \item \textbf{Headpose:}
    The head movements along the x, y, and z axes are not uniform or consistent in a deepfake compared to a real video. A tendency to move our heads while talking is common; however, upon conducting research, two prominent studies by \cite{b30} and \cite{b31} that this is not the case for deepfakes. Often in a deepfake, the head movement does not represent the natural face as it looks more like a mask. This leads to irregular movements where even though the head is completely on the other side, the face is not. Hence, head movement can be used as a parameter to gauge whether the video is a deepfake or not. The camera matrix was utilized to gain depth perception in the frame. The solvePnP function  was used to get the rotational and translational vectors that describe the 3D pose of the face relative to the camera. Then cv2’s Rodrigues transformation was used to convert the rotational vector into a rotational matrix, following which the RQDecomp3x3 function was used to convert the rotational matrix into Euler angles which were then used to calculate and estimate the headpose.
    \begin{figure}[H]
    \centering
    \includegraphics[width=10cm]{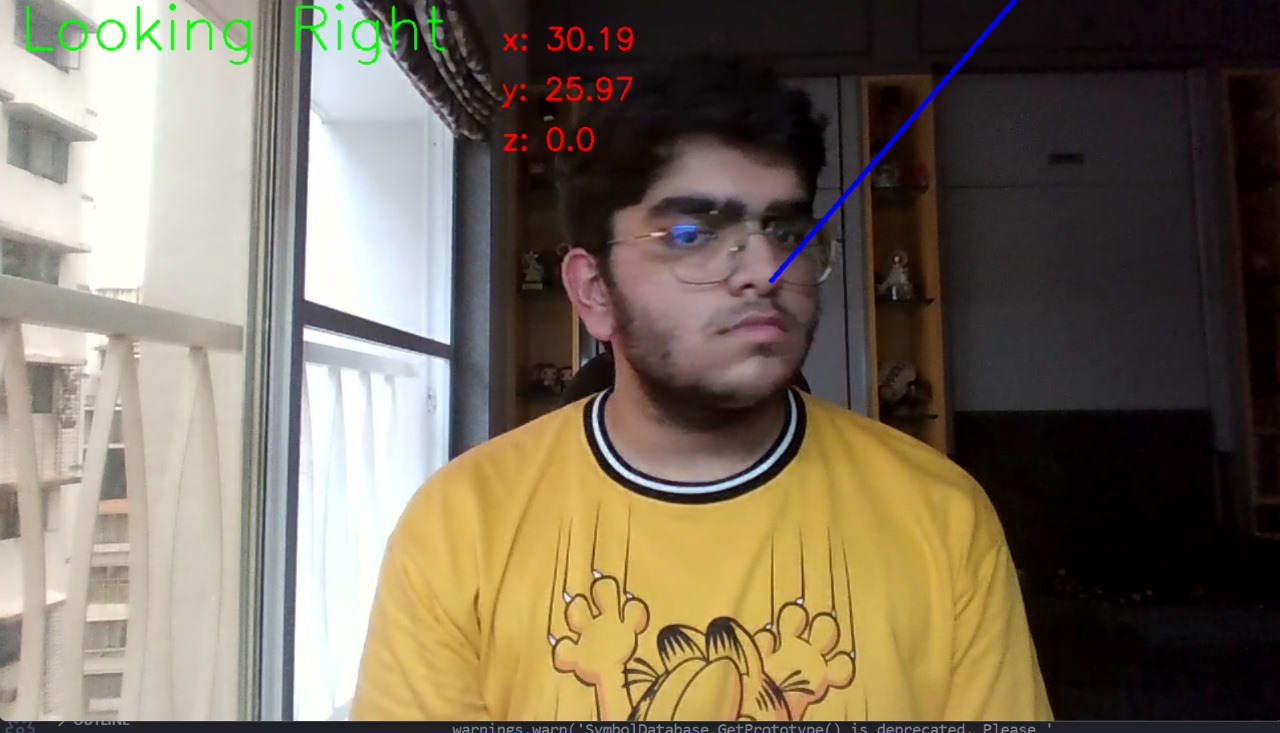}
    \caption{\label{fig:equation1}Headpose estimation, using camera matrix and 3D pose calculations to analyze head movement inconsistencies.}
    \end{figure}

    \item \textbf{Skin Tone:}
    Skin tone helps in facial color analysis by detecting changes in blood flow and its concentration. It has been proven that color spaces other than RGB perform better while detecting color according to Hadas, Shahar et al \cite{b32}. Hence, use of the oRGB color space was made. The oRGB color space is a color space that separates color information into different channels to enhance features for skin detection. It models color perception using three channels:
    \begin{enumerate}
    \item Luminance [L]: A grayscale value representing brightness.
    \item Chrominance [C1]: The difference between red and green color channels.
    \item Chrominance [C2]: The difference between blue and yellow colors.
\end{enumerate}
    To convert an RGB pixel to an oRGB pixel, the following linear transformation is performed:
    \[
    \begin{bmatrix}
    L \\
    C_{1} \\
    C_{2}
    \end{bmatrix}
    =
    \begin{bmatrix}
    0.299 & 0.587 & 0.114 \\
    0.500 & 0.500 & -1.000 \\
    0.866 & -0.866 & 0.000
    \end{bmatrix}
    \begin{bmatrix}
    R \\
    G \\
    B
    \end{bmatrix}
    \]
\end{itemize}




\begin{figure}[H]
\centering
\includegraphics[width=12cm]{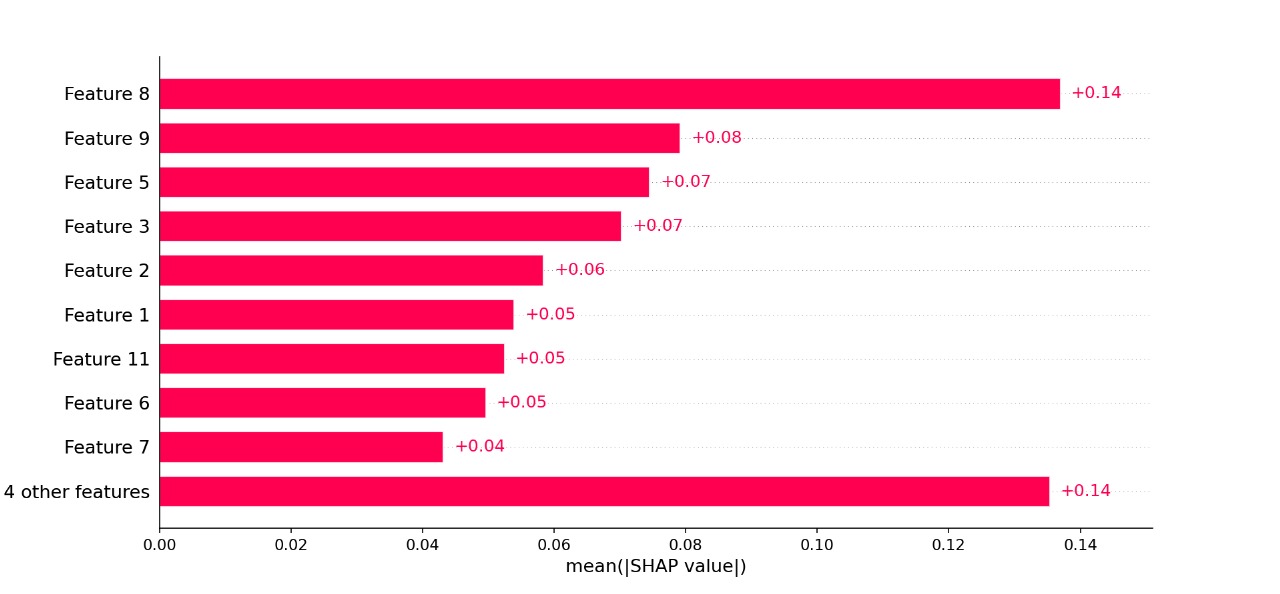}
\caption{\label{fig:feature_importance}Feature importance for deepfake video detection. The x-axis represents the mean absolute SHAP (SHapley Additive exPlanations) values, indicating the average impact of each feature on the model's output. The features include cheekbone height, inter-pupil distance, number of blinks, headpose angles (x, y, z), nose size, lip size, contrast correlation, luminance, chrominance1, chrominance2, and others, listed from 1 to 13, respectively. Higher SHAP values indicate greater importance of a feature in the model's predictions.}
\end{figure}

\subsubsection{4.1.3 For Audio}
To understand and visually study audio signals, mel-spectrograms were utilized. ”Mel” is an abbreviation for ”melody.” Mel-spectrograms \cite{b27} provide a time-frequency representation of audio signals with perceptually relevant amplitude and frequency representations. Both amplitude and frequency perceptions are nonlinear and can be expressed in logarithmic form. The mel scale is derived from a perceptually informed scale for the pitch of sound. The pitch for a 1kHz  frequency of sound is perceptually similar to 1000 mels, making the pitch of the sound equivalent. The studies in \cite{b33}, \cite{b34}, and \cite{b35} provided us with a deep understanding of mel-spectrograms, which was effectively applied in our research.

The formula was from \cite{b34} to conduct the following:
\begin{equation}
    m = 2595 \cdot \log_{10} \left(1 + \frac{f}{700} \right)
\end{equation} 
\begin{equation}
    f = 700 \left(10^{\frac{m}{2595}} - 1 \right)
\end{equation}


\textbf{Steps for Mel-Spectrogram Generation}

\begin{enumerate}[label=\textbf{\arabic*}]
    \item Perform Short-Time Fourier Transform (STFT)
    \item Convert Amplitude to Decibels (dB)
    \item Convert Frequencies to Mel Frequency Representation
\end{enumerate}

\begin{figure}[H]
\centering
\includegraphics[width=12cm]{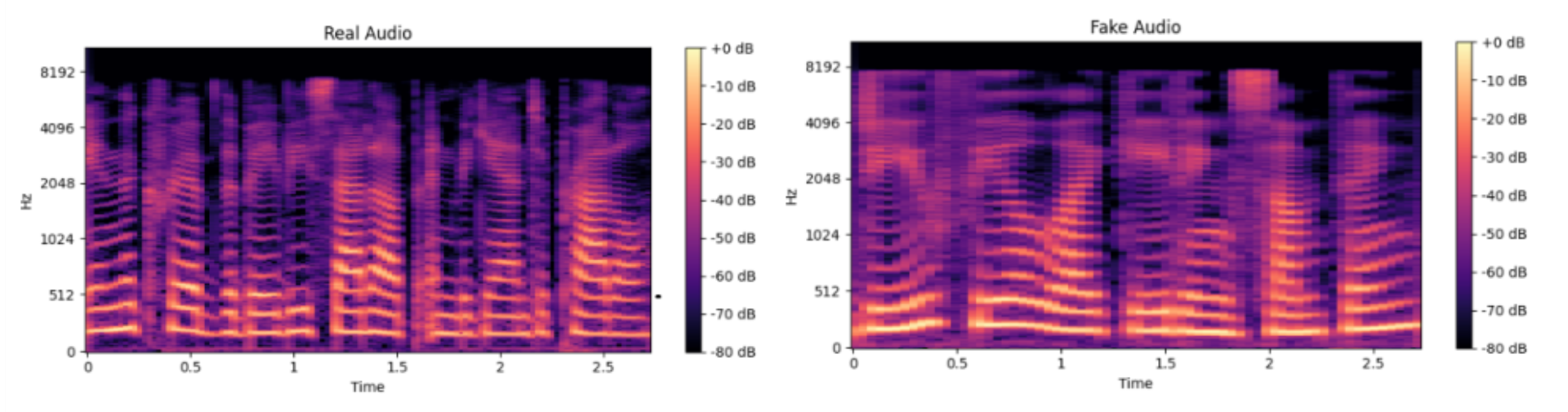}
\caption{\label{Figure 1}Mel-spectrograms comparing real (left) and deepfake (right) audio signals reveal distinct differences in time-frequency representation and amplitude. The fake audio often exhibits a broader frequency range and unique spectral signatures, with more harmonics and clearer patterns, unlike the real audio, which includes background noise and vocal imperfections.}

\end{figure}

\textbf{Choosing the Appropriate Number of Mel Bands}\\
The number of mel bands depends on the specific problem. For the research, 128 bands were used. This choice is informed by the 88 notes on a piano, which correspond to approximately 90 mel bands, aligning with the notes of Western music state in \cite{b18}.\\

\textbf{Construction of Mel Filter Banks}
\begin{enumerate}[label=\textbf{\arabic*}]
    \item Convert the lowest and highest frequencies to their mel representations using the formula for \( m \).
    \item Create bands with equally spaced points, based on the desired number of mel bands.
    \item Convert these points back to Hertz.
    \item he frequency bins were rounded to the nearest value due to the discrete nature of signals and the constrained resolution imposed by the frame size of the Short-Time Fourier Transform (STFT).
    \item Create triangular filters, which are the building blocks of the mel scale.
    \item Higher frequencies have larger gaps between points to achieve the same pitch compared to mel frequencies, which have similar pitch.
    \item When plotted, this shows triangular-shaped filters.
    \item The shape of mel filter banks is geometric, but the calculations are algebraic.
\end{enumerate}

\begin{figure}[H]
\centering
\includegraphics[width=10cm]{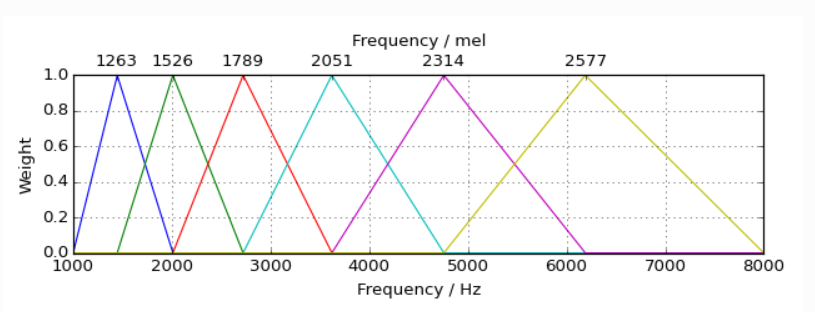}
\caption{\label{Figure 1}Plot of mel filter bank weights against mel frequencies and Hertz frequencies. The graph visualizes how triangular mel filters map frequency bands from Hertz to the mel scale, demonstrating the mel scale's frequency distribution \cite{b36}.}
\end{figure}

\textbf{Applying mel filter Bank to Matrix to Normal spectrograms}
\begin{equation}
    M_{\text{MelFilterBank}} = (\text{number of bands}, (\text{frame size} / 2) + 1)
\end{equation}

\begin{equation}
    Y_{\text{Spectrogram}} = ((\text{frame size} / 2) + 1, \text{number of frames})
\end{equation}

\begin{equation}
   Mel-spectogram= M \cdot Y_(\text{number of bands}, \text{number of frames})
\end{equation}
Matrix multiplication is possible as the rows of \( M \) and \( Y \) are the same.

Each point in the graph indicates the presence of a mel band at a specific point in time. This is represented using different color combinations based on dB values. Mel spectrograms find applications in various fields such as audio classification, music genre classification, music instrument classification, and automatic mood recognition systems. 

\subsection{Models Used}
\subsubsection{4.2.1 For Video}
Initially, the videos  were processed through our feature extraction model, resulting in a final feature dataframe with 2,590x13. Given the high imbalance between fake and real videos, the SMOTE (Synthetic Minority Over-sampling Technique) method was employed to upsample the dataset. After upsampling, the feature dataframe expanded to 4,342x13. The data was subsequently split into training and testing sets in a 80:20 ratio. This was then fed to the various models mentioned below.

Decision Trees were used to identify deepfake videos because it effectively handles various features extracted from videos, such as facial landmarks, texture characteristics, and skin tone. However, they fail with high-dimensional and complex data which might lead to overfitting and poor generalization. To tackle this, Random Forest was used to identify deepfake videos because of its ability to improve classification accuracy and robustness through ensemble learning. By constructing multiple decision trees and pooling their predictions, random forests reduce overfitting and increase generalization. This approach efficiently handles diversity and complexity from video data, combining the strengths of individual trees to provide more reliable classification.

Since this approach did not effectively capture intricate patterns and temporal dependencies, bagging was employed to enhance model stability and accuracy by reducing variance. By training multiple models on different bootstrap samples of the data and comparing their predictions, bagging reduces the risk of overfitting any one sample. To increase the accuracy, XGBoost was used as it focuses on patterns that are difficult to classify. This method iteratively refines the model by emphasizing misclassified instances, leading to a more robust and accurate detection system for distinguishing between real and deepfake videos.

All the above methods used machine learning techniques which can't handle complex patterns as effectively as deep learning models. With the intention of creating our own deep learning model and not using the pre-trained ones, Artifical Neural Network (ANN) for DeepFake video detection was chosen as they excel in recognizing complex patterns and representations from data. Inspired from the human brain, ANNs consist of multiple neural networks that learn certain features through adaptive training. This capability is crucial for detecting subtle and complex differences in real and deepfake video.

The feedforward structure of ANNs allows them to process input data across multiple layers, capturing nonlinear relationships and high abstractions that are missing in simple models. The use of activation functions helps detect nonlinearities, enabling the network to model complex patterns. Furthermore, the binary cross-entropy loss function efficiently handles the classification task by considering the difference between the predicted probabilities and the actual labels, which guides the network to improve the accuracy of the difference between real and deepfake video.

For a feedforward neural network, the output of a single neuron \( j \) in layer \( l \) is:

        \[
        a_j^{(l)} = \sigma\left( \sum_{i=1}^{n_{l-1}} w_{ij}^{(l-1)} a_i^{(l-1)} + b_j^{(l)} \right)
        \]

        Where \( \sigma \) is the activation function, \( w_{ij} \) are the weights, \( a_i \) are the activations from the previous layer, and \( b_j \) is the bias term.
  \vspace{15px}

 \textbf{Loss Function}
        For binary classification, the loss function (binary cross-entropy) is:

        \[
        L = - \frac{1}{N} \sum_{i=1}^{N} \left[ y_i \log(\hat{y}_i) + (1 - y_i) \log(1 - \hat{y}_i) \right]
        \]

        Where \( y_i \) is the true label, \( \hat{y}_i \) is the predicted probability, and \( N \) is the number of samples.
        \begin{figure}[H]
        \centering
        \includegraphics[width=12cm]{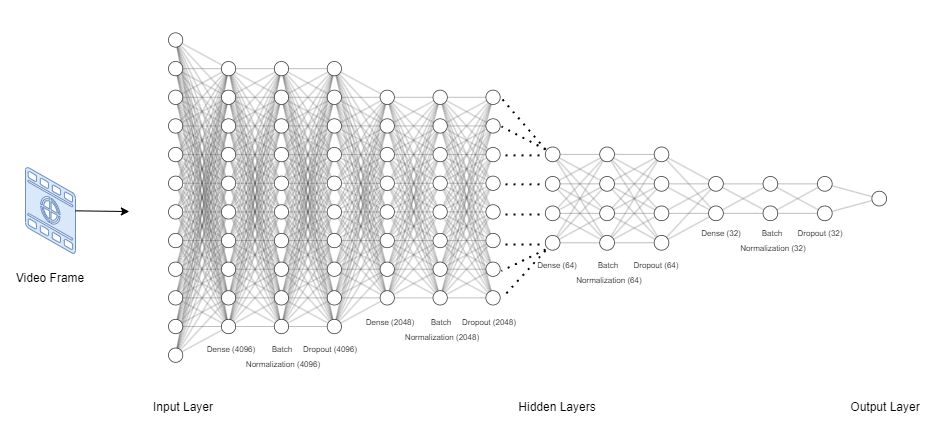}
        \caption{\label{fig:equation1}Architecture of the Artificial Neural Network (ANN) used for DeepFake video detection, illustrating the feedforward structure with multiple layers and activation functions to capture complex patterns.}
        \end{figure}


\subsubsection{4.2.2 For Audio}
After extracting the features split the data into training-testing ration of 80:20.
During training, validation data comprising a randomly distributed set of samples was used to monitor the model's performance.

    Random Forest and XGBoost were used to classify labeled audio samples due to their strong performance in complex data processing. Random forest with a sufficient number of estimators effectively reduces overfitting and provides robust classification results. XGBoost provided increased flexibility with its wide range of parameters, allowing to fine-tune the model for improved accuracy. Despite achieving adequate accuracy with these techniques, deep learning methods were selected to further reduce loss and enhance model performance, leveraging their capability to capture complex patterns in audio data.

    Convolutional Neural Networks (CNNs) were used for classifying mel spectrograms because CNNs excel at identifying and learning spatial hierarchies in image-like data. In our case, mel-spectrograms are 2D representations of audio signals, where spatial patterns can reveal important features for classification. CNNs are particularly effective at detecting these patterns through their convolutional layers, which apply filters to capture features such as edges, textures, and shapes.

The architecture of our CNN, with layers like Conv2D and MaxPooling2D, is designed to extract and downsample features from the mel-spectrograms, creating feature maps that highlight relevant information. The use of layers like Flatten, Dense, and Dropout further helps in combining these features, preventing overfitting, and improving classification performance. By leveraging CNNs, complex patterns were easily learned and achieved higher accuracy in distinguishing between different audio samples.
 \begin{figure}[H]
        \centering
        \includegraphics[width=10cm]{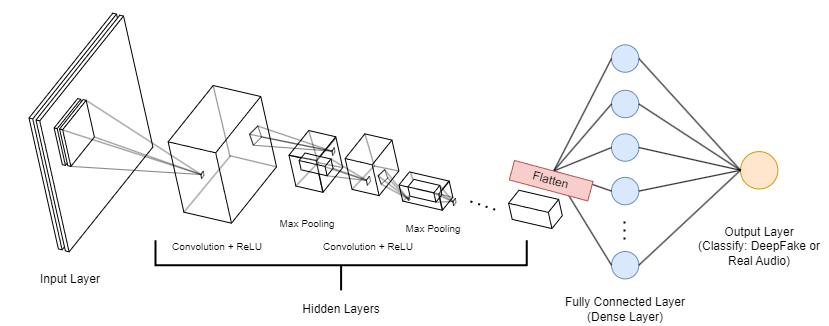}
        \caption{\label{fig:equation1}Architecture of the Convolutional Neural Network (CNN) used for classifying mel-spectrograms, highlighting layers such as Conv2D and MaxPooling2D for feature extraction and classification.}
\end{figure}

   VGG19 was used to achieve the highest accuracy for our classification task due to its proven performance and the advantages of transfer learning. VGG19 is a well-established convolutional neural network architecture known for its deep and uniform layer structure, which allows it to capture intricate features from images. By utilizing pre-trained weights from ImageNet, VGG19 provides a strong starting point with learned features that can be adapted to our specific problem.

VGG19 was configured with an input shape of 224×224×3 to match the size and color channels of our mel-spectrogram images. By freezing all but the last 4 layers of the pre-trained VGG19 model, the valuable feature extraction capabilities were preserved while allowing fine-tuning on our specific dataset. This approach leverages the robust features learned from a large, diverse dataset while adapting the model to our task. The inclusion of additional dense and dropout layers refined the model and mitigated overfitting, ultimately boosting its performance for binary classification.

 \begin{figure}[H]
        \centering
        \includegraphics[width=12cm]{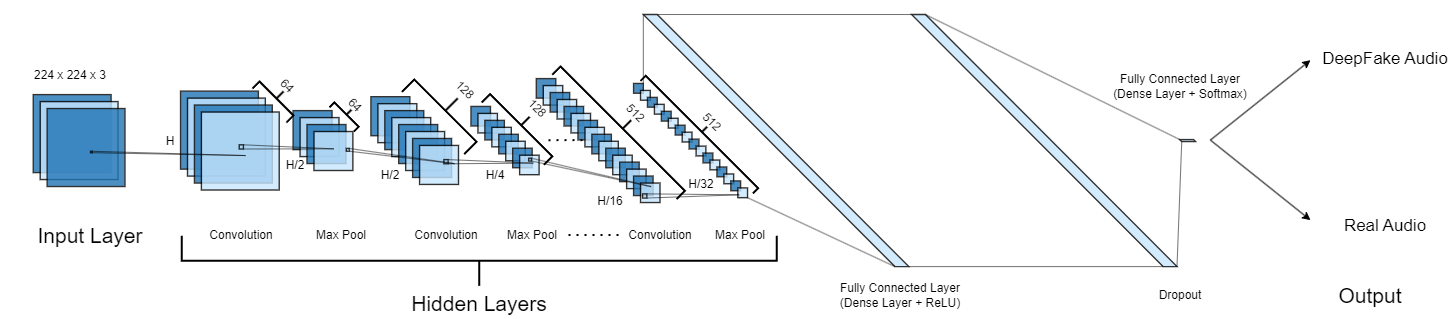}
        \caption{\label{fig:equation1}VGG19 architecture adapted for mel-spectrogram classification, showcasing transfer learning with pre-trained weights and fine-tuning for enhanced performance.}
\end{figure}

 \section{Experimental setup}
 The devices used for data collection and video feature extraction contained processors using i5 12th generation with Nvidia rtx 3050 16 gb ram. Similarly devices used for audio feature extraction contained processors using i7 12th generation with Nvidia rtx 3060, ddr6 gpu. 
 
\section{Results}
The model's performance is summarised in a classification report, which includes metrics such as precision, recall, F1 score, and accuracy. Table 1 presents a detailed classification report for DeepFake video detection and Table 2 presents the detailed classification report for DeepFake Audio detection 







\begin{table}[h!]
\centering
\caption{Performance Metrics for Various Methods for DeepFake Video}
\begin{tabular}{lcccc}
\toprule
\textbf{Method} & \textbf{Precision} & \textbf{Recall} & \textbf{F1-Score} & \textbf{Accuracy} \\
\midrule
Decision Tree & 0.80 & 0.80 & 0.80 & 0.80 \\
Random Forest & 0.88 & 0.88 & 0.89 & 0.89 \\
Bagging & 0.94 & 0.90 & 0.92 & 0.92 \\
XGBoost & 0.94 & 0.91 & 0.92 & 0.92 \\

\textbf{ANN} & \textbf{0.88} &\textbf{0.96} & \textbf{0.93} & \textbf{0.93} \\
\bottomrule
\end{tabular}
\label{tab:performance_metrics}
\end{table}






\begin{table}[h!]
\centering
\caption{Performance Metrics for Various Methods for DeepFake Audio}
\begin{tabular}{lcccc}
\toprule
\textbf{Method} & \textbf{Precision} & \textbf{Recall} & \textbf{F1-Score} & \textbf{Accuracy} \\
\midrule
Random Forest & 0.83 & 0.82 & 0.82 & 0.82 \\
Gradient Boosting & 0.86 & 0.86 & 0.86 & 0.86 \\
CNN & 0.90 & 0.91 & 0.90 & 0.90 
\\
\textbf{VGG19} & \textbf{0.98} & \textbf{0.97} & \textbf{0.98} & \textbf{0.98} \\
\bottomrule
\end{tabular}
\label{tab:performance_metrics_audio}
\end{table}





 \begin{figure}[!ht]
        \centering
        \includegraphics[width=12cm]{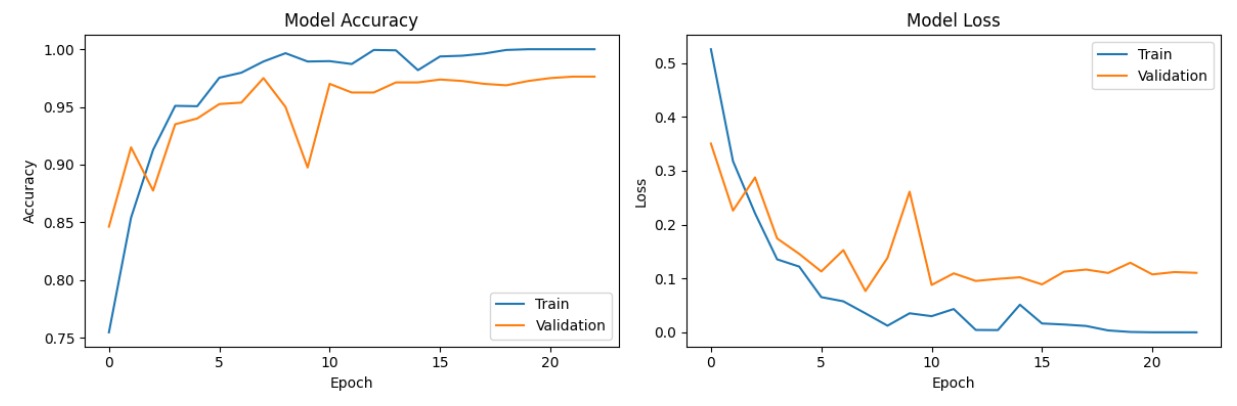}
        \caption{\label{fig:equation1}Training performance of the VGG19 model, displaying accuracy (left) and loss (right) versus epochs, illustrating the model's learning progress over time.}
\end{figure}

\textbf{Proposed Multimodal Approach Results}:
The original dataset was utilized in splits for training and testing, while randomly swapping real audio with deepfake audio. During testing, a balanced distribution of samples is ensured across the following categories: 'real-deepfake' (real video with deepfake audio), 'deepfake-real' (deepfake video with real audio), 'real-real' (real video with real audio), and 'deepfake-deepfake' (deepfake video with deepfake audio). The models with the highest accuracy were used for video and audio classification to generate the final output for the video.If either the video or audio component is identified as deepfake, the overall sample is classified as deepfake. \\
Table 3 shows that the approach correctly classified 1955 samples out of 2079 samples giving it a accuracy of \textbf{94\%}. 

\begin{table}[h!]
\centering
\caption{Correctly Classified Data for Multimodal Data}
\begin{tabular}{cccc}
\toprule
\textbf{Video} & \textbf{Audio} & \textbf{Number of Samples} & \textbf{Correctly Classified} \\
\midrule
0  & 0  & 528 & 502  \\
0 & 1  & 523 & 496  \\
1 & 0 & 513 & 477  \\
1  & 1  & 515 & 480  \\
\bottomrule
\end{tabular}
\label{tab:performance_metrics_audio_video}
\end{table}
Here, '0' denotes real media whilst '1' denotes deepfake media.

 \begin{figure}
\begin{subfigure}{0.9\linewidth}
    \centering
    \includegraphics[width=\linewidth]{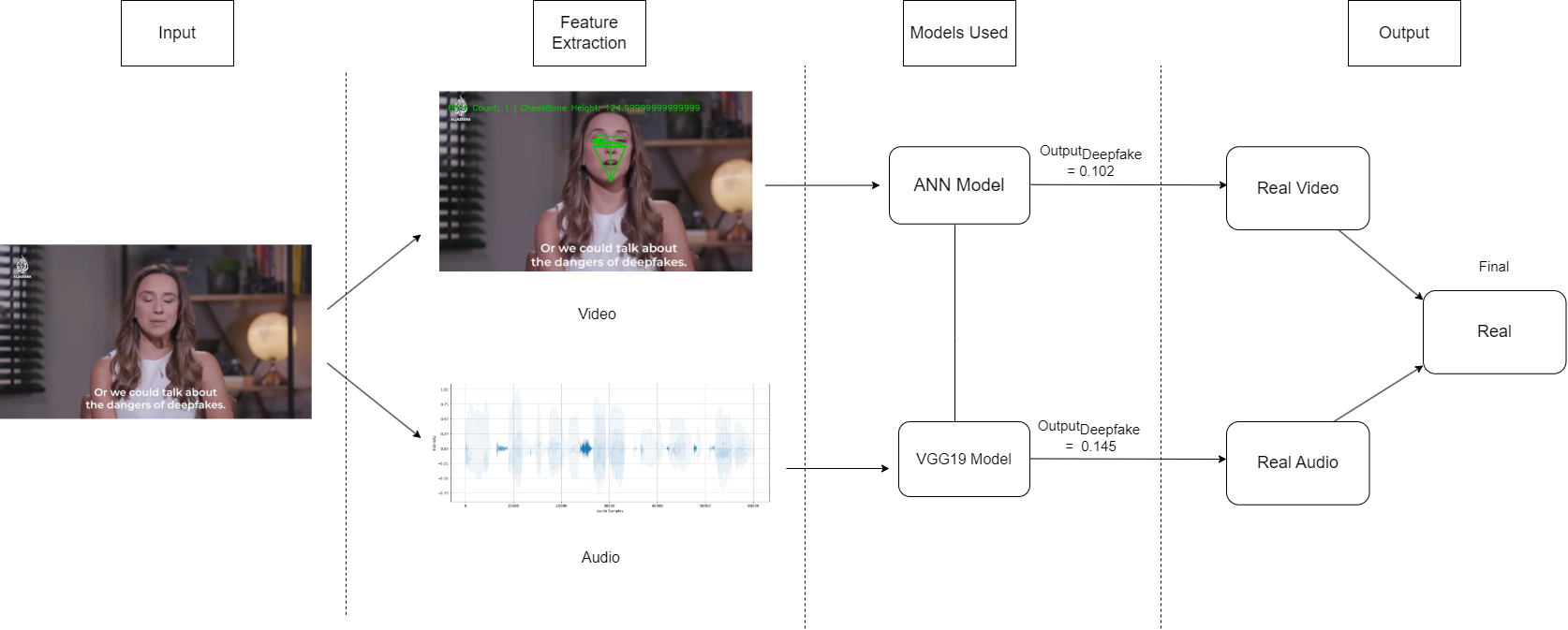}
    \caption{Classification results for 'Real-Real' samples, showing an overall classification of 'Real.'}
    \label{fig:real_real}
\end{subfigure}

\begin{subfigure}{0.9\linewidth}
    \centering
    \includegraphics[width=\linewidth]{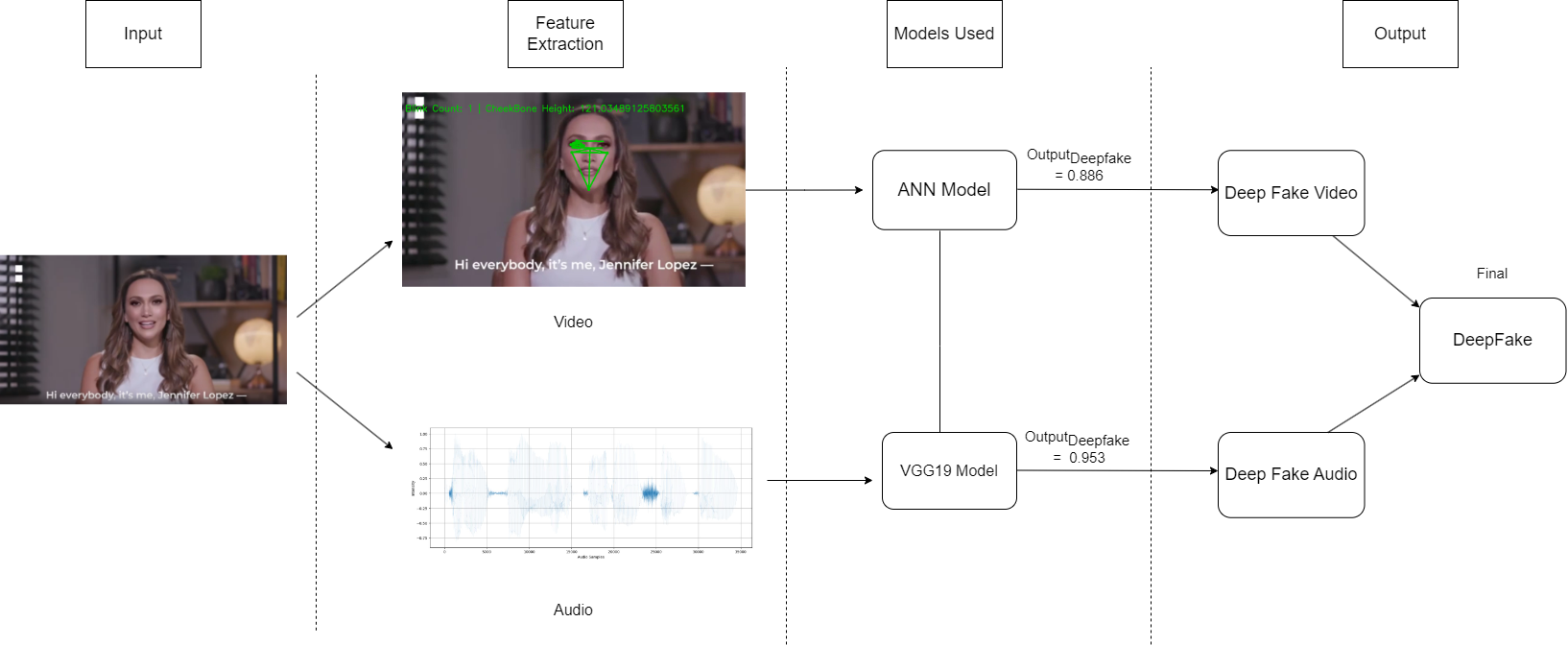}
    \caption{Classification results for 'DeepFake-DeepFake' samples, showing an overall classification of 'DeepFake.'}
    \label{fig:deepfake_deepfake}
\end{subfigure}
\begin{subfigure}{0.9\linewidth}
    \centering
    \includegraphics[width=\linewidth]{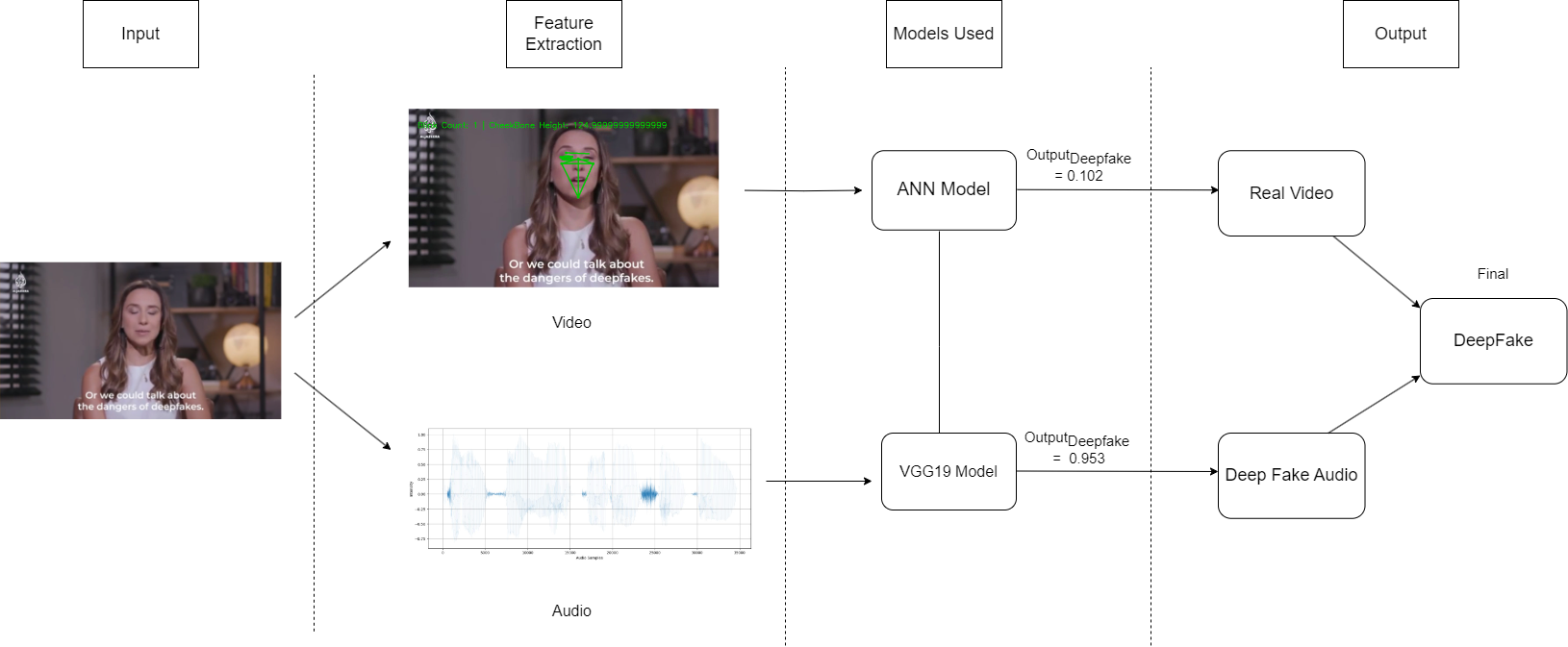}
    \caption{Classification results for 'Real-DeepFake' samples, showing an overall classification of 'DeepFake.'}
    \label{fig:real_deepfake}
\end{subfigure}

\begin{subfigure}{0.9\linewidth}
    \centering
    \includegraphics[width=\linewidth]{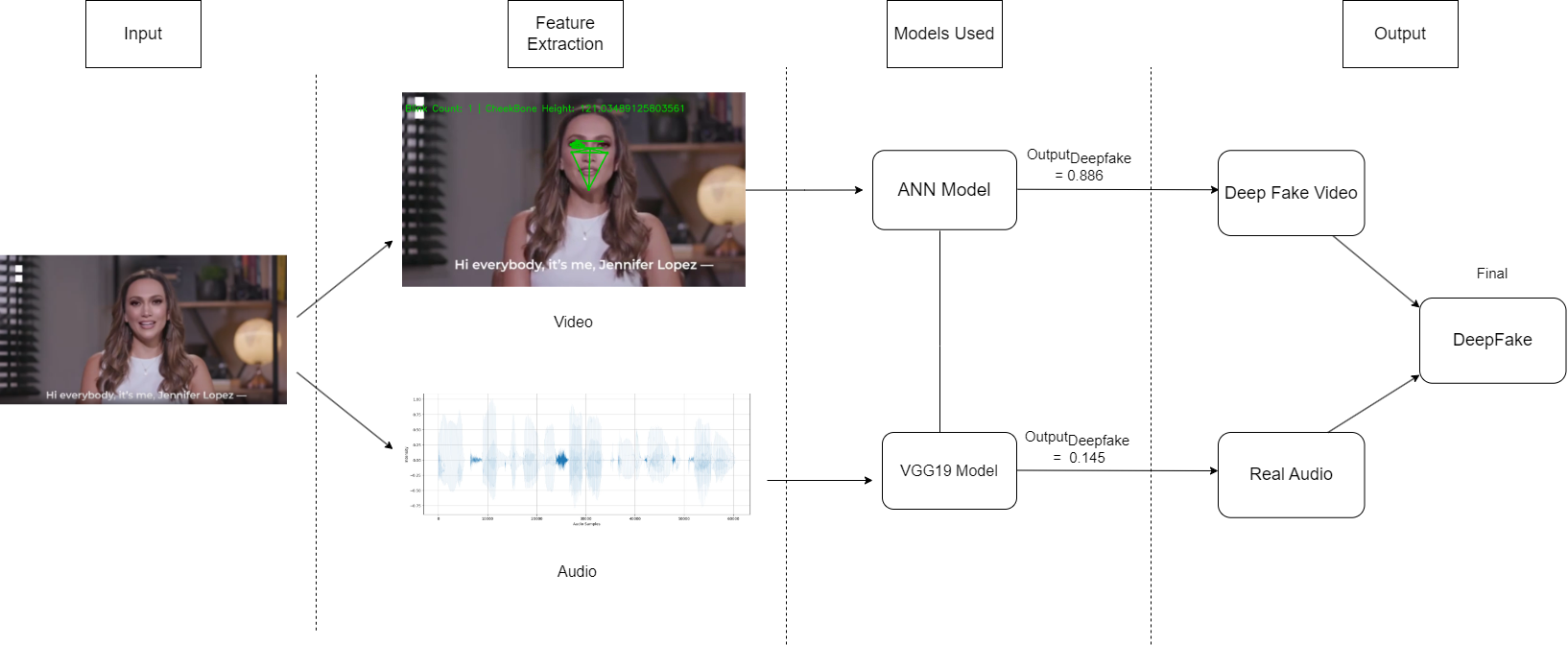}
    \caption{Classification results for 'DeepFake-Real' samples, showing an overall classification of 'DeepFake.'}
    \label{fig:deepfake_real}
\end{subfigure}
\label{fig:combined_results}
\end{figure}

\section{Conclusion}
The deepfake detection methodology employs advanced deep learning techniques for both video and audio analysis. Specifically, ANN is utilized for video classification with an accuracy of 93\%, while transfer learning with VGG19 is used for audio classification, achieving an accuracy of 98\%. These models outperform traditional algorithms like Random Forest, Decision Tree, XG-Boost, and Bagging, which were less effective in the comparative analysis. The combined approach results in an overall accuracy of 94\%, demonstrating its robustness and effectiveness.
Our approach improves on previous works, such as the 86.13\% accuracy with XG-Boost in \cite{b6}, by integrating multiple visual features into a comprehensive multimodal framework. While studies like \cite{b29}, \cite{b30}, and \cite{b32} focus on single features—such as textural features or headposes—they often lack reliability on unseen data where other features may vary. By leveraging multiple features, our method offers greater robustness in detecting deepfakes across diverse scenarios.

In contrast, studies like \cite{b8}, \cite{b9}, and \cite{b10} face challenges due to computational constraints and model inefficiencies. For example, \cite{b8} combined resource-intensive models, leading to high computational costs, while \cite{b9} used a 3D CNN with high memory usage. Despite using multiple models, \cite{b10} achieved only 77\% accuracy, underscoring the need for a more streamlined and effective approach like ours.

The detection of deep fake audio in our work surpasses previous methods, such as in \cite{b33}, where a custom CNN achieved 88.9\% top-5 accuracy and VGG19 reached 88.5\%, both significantly lower than our model's 98.0\%. Similarly, \cite{b19} implemented a 7-layered CNN with 91\% accuracy using MFCC features, but our model outperformed it without needing MFCC extraction, enabling faster processing. Additionally, they lacked real-time evaluation results, which our model successfully achieved.

By combining visual and audio features, our multimodal approach addresses the limitations of unimodal systems, ensuring comprehensive feature extraction and robust detection. This method demonstrates superior accuracy and reliability, making it a significant advancement in the field of deepfake detection.

\section{Future Scope}

The DFDC dataset for deepfake video detection had 50 zip files, but our model was trained on just 1 zip file with 3,135 videos. With enhanced computational resources, this approach could be extended to the entire dataset and other datasets, improving deepfake detection and preventing the spread of fake media.

For audio detection, future improvements could include real-time audio monitoring alongside visuals. Our models were trained on a reduced Fake-or-Real dataset with 4,000 audio samples. Better extraction methods and data creation with appropriate dimensions could enhance performance, as computational time for optimal input files was higher than expected.

In combined analysis, the real and deepfake video and audio were swapped to create test data. A more comprehensive dataset with balanced samples of all combinations (deepfake-real, real-deepfake, real-real, deepfake-deepfake) would improve evaluation accuracy and refinement of our multimodal approach.\\

\textbf{Acknowledgement}
We would like to express our deepest gratitude to our mentor, Chaitya Shah [chaitya0623@gmail.com], for his unwavering support and insightful feedback throughout this research. His guidance has been instrumental in refining the methodology and scope of the study, significantly contributing to the development and success of this work.

\end{document}